\documentclass[a4paper,conference]{IEEEtran}
\usepackage{cite}

\ifCLASSINFOpdf
   \usepackage[pdftex]{graphicx}
\else
\fi

\usepackage{subcaption}
\usepackage{multirow}
\begin{document}

%
\title{Weakly Supervised Scene Text Detection using Deep Reinforcement Learning}

\author{\IEEEauthorblockN{Emanuel Metzenthin, Christian Bartz, Christoph Meinel}
\IEEEauthorblockA{emanuel.metzenthin$@$student.hpi.de, \{christian.bartz, christoph.meinel\}$@$hpi.de \\
Hasso Plattner Institute, University of Potsdam\\
Prof.-Dr.-Helmert Strasse 2-3 \\
14482 Potsdam, Germany}}


%


\maketitle

\begin{abstract}
The challenging field of scene text detection requires complex data annotation, which is time-consuming and expensive. Techniques, such as weak supervision, can reduce the amount of data needed. In this paper we propose a weak supervision method for scene text detection, which makes use of reinforcement learning (RL). The reward received by the RL agent is estimated by a neural network, instead of being inferred from ground-truth labels. 
First, we enhance an existing supervised RL approach to text detection with several training optimizations, allowing us to close the performance gap to regression-based algorithms.
We then use our proposed system in a weakly- and semi-supervised training on real-world data. Our results show that training in a weakly supervised setting is feasible. However, we find that using our model in a semi-supervised setting , e.g. when combining labeled synthetic data with unannotated real-world data, produces the best results.
\end{abstract}

%
\IEEEpeerreviewmaketitle

\section{Introduction}

In the past years we have seen a tremendous technical advancement in computer vision algorithms \cite{He, Redmon, UjjalDey}. But this success does not come without a cost, namely the need for large amounts of labeled training data \cite{Yang}. Among computer vision tasks, annotation for text detection is one of the most complex forms, as it requires bounding box labels to mark the exact text locations in the image.
According to \cite{Zhao} it takes on average 39 seconds to label images containing multiple texts with tight bounding boxes, a 20-fold increase over annotation for classification. 

Scene Text Detection (STD), is known as the method of determining
the locations of text in natural scene images \cite{Cao2020}, e.g., on storefronts or street signs, often called "text in the wild" \cite{Khan2021}.

Its use cases include content-based image search, automatic navigation (including robotics) \cite{Lin2020},  instant translation \cite{Wang} as well as the digitization of document archives \cite{Lucas2005}. In most cases STD is used as a preliminary step for text recognition \cite{Lin2020}. Narrowing down the search space in natural images and proposing accurate text regions is crucial for recognition algorithms to perform well.

STD is a notoriously hard problem. One must distinguish it from the much more structured problem of optical character recognition (OCR) in scanned documents \cite{Lin2020}. In contrast to document text, scene text has no underlying formatting, can occur in a variety of sizes, fonts, colors and orientations and be scattered across the image \cite{Lin2020, Yin}. Real-world scenes also come with very diverse backgrounds, requiring large amounts of data for algorithms to learn from \cite{Lin2020}. For all specialized applications, for which public datasets or pretrained models do not suffice, this data has to be manually annotated. 

Many supervised STD methods have been developed in the past (see \ref{sec:rel_work}). Most of these approaches follow a bottom-up approach using region proposals \cite{Kong}. Only some top-down methods involving reinforcement learning (RL) have been tested. This approach not only forms an alternative to the standard route taken by most researchers, which reduces the number of region proposals required \cite{Caicedo2015}, but also brings the advantage that it relies on a single scalar reward. This reward can be estimated and thus opens the door to weaker forms of supervision. 

We combine this idea with a neural network introduced by Bartz et al. \cite{Bartz}, which rates the RL detections (see \ref{sec:weak_sup}). 
Weak supervision is a paradigm where noisy labels are applied to otherwise unlabeled data points. These labels are called \textit{weak} as they are not guaranteed to be ground-truth, but come from heuristics or estimates. When the model is taught properly what good text detections look like, it can serve as an automated annotator for unlabeled images. 
Our experimental results with this model can be found in \ref{sec:results} and suggested future improvements in \ref{sec:discussion}. We release our code and models to the community \footnote{https://github.com/emanuel-metzenthin/RL-scene-text-detection}.\\

We make the following contributions in this paper:

\begin{enumerate}
    \item We enhance an existing supervised RL approach to STD and show contrary to prior research that high detection rates are possible using this method.
    \item We combine the weak supervision technique for object detection proposed by \cite{Bartz} with our RL method and show that weakly supervised STD is feasible this way.
    \item We perform semi-supervision using synthetic labeled data and weakly supervised real-world data
\end{enumerate}

\section{Related Work}
\label{sec:rel_work}
Recently, deep learning methods have dominated the field of STD, most of which can be categorized as either bounding box regression or segmentation-based methods \cite{Khan2021}. Only few RL methods have been proposed for visual detection tasks. Some work has been done on weakly supervised STD, mainly by encoding complex shape information in simpler annotation formats.

\subsection{Bounding Box Regression Methods}

Bounding box regression methods primarily adapt models from the general object detection domain and rely on a region proposal stage which produces boxes likely to contain objects. Most regression-based text detection approaches build on top of region proposal networks (RPNs) \cite{Ren}. RPNs slide windows based on predefined anchor boxes over a CNN feature map \cite{Ma}. A classification branch then determines for each candidate box whether it contains an object and fine-tunes the detection bounding box location.

An example of a RPN-based method is \textit{Deep Text} by Zhong et al. \cite{Zhong2016}.
Ma et al. \cite{Ma} and Jiang et al. (\textit{R2CNN}) \cite{Jiang2017} also rely on a RPN, but adapt the region proposals to include a rotation dimension.

Tian et al. \cite{Tian2016} approach the problem by generating very small fixed-sized candidate regions and determine whether they show text characteristics. These small regions are then combined to text lines. Their model is named \textit{Connectionist Text Proposal Network (CTPN)}.

\subsection{Segmentation Methods}

Image segmentation methods usually rely on fully convolutional networks (FCN) \cite{Khan2021}. These networks contain a downsampling and upsampling convolution pass and produce a pixel-wise prediction map of text/no-text areas in an image. Due to the fact that FCNs produce a more fine-grained output than region proposals based on bounding boxes, segmentation based methods have become mainstream for detecting complex arbitrarily shaped text \cite{Cao2020}. 

He et al. \cite{He2016} propose the \textit{Cascaded Convolutional Text Networks} (CCTN) which consists of two CNNs. A coarse network outputs a heatmap for probable text regions which get fine-tuned into accurate segmentations in a second stage.

Long et al. \cite{Long2018} propose the \textit{TextSnake} algorithm to allow for the detection of irregular text shapes. They model the text geometry using disks with varying radii slid along the center line of a text instance. Their FCN produces a segmentation map for text regions and text center lines. The text region map is used to mask the center line prediction and eliminate false positives. 

The \textit{Progressive Scale Expansion Network} (PSENet) by Wang et al. \cite{Wangb} produces several segmentation masks at different scales. These are then combined to form more accurate detection masks.

Segmentation-based methods often incorporate the entire image information into their calculations and can therefore be superior in complex scenes \cite{Khan2021}. On the other hand, they are usually slower than regression-based techniques and often fail to segment overlapping text \cite{Khan2021, Cao2020}.

\subsection{Reinforcement Learning Methods}

Only a few works have been published that make use of RL as the main algorithm component for detection. They mostly target object detection but can often be transferred to text detection as well. Most of the methods aim to build a faster top-down alternative to the bottom-up region proposal based object detection, which often requires the analysis of thousands of regions per image \cite{Kong}. 

Caicedo et al. \cite{Caicedo2015} were the first to develop such an active search detector and also laid the groundwork for the first RL text detection algorithm proposed by Wang et al. \cite{Wang2018} as well as for our algorithm design. \\
They use deep Q-learning to learn a RL policy that manipulates a bounding box on top of an image until it covers the sought-after object. 

A similar approach was followed by Mathe et al. \cite{Mathe}. However, their algorithm uses a stochastic policy that decides on fixation points in an image, instead of transforming a bounding box.

Kong et al. \cite{Kong} improve Caicedo et al.'s work by adapting their approach to a multi-agent collaborative RL setting. Instead of a single, multiple agents search the image for objects, trained jointly via gated cross connections. 

Bueno et al. \cite{Bellver2016} also try to improve upon \cite{Caicedo2015}. Before their RL agent begins its visual search the image is zoomed into one of 5 predefined regions. This hierarchical image search allows for detection with fewer steps.

Another variant of object detection RL was introduced by Jie et al. \cite{Jie}. They base their approach on a tree-search. Starting with the entire image as the root node of a tree, a scaling and a translation action are chosen at each tree level. The leaf node representing the best bounding box is then selected as the final detection. 

\subsection{Weak Supervision Methods}

Weak supervision in STD has not been studied extensively, but some works have been published that aim to reduce the cost associated with data annotation. Most of them try to replace polygon annotations for complex text shapes with relaxed annotation formats. Others experiment with semi-supervision.

Zhuang et al. \cite{Zhuang2021} develop an annotation tool that facilitates the annotation of polygonal bounding boxes. The annotator only selects points along the center line of text instances. A text recognizer is then applied and the boundary producing the lowest recognition loss is selected as the final label.

Zhao et al. \cite{Zhao} develop a text detector which predicts pseudo labels for candidate proposals. This pseudo label generation is learned based on weak labels of different granularities. The authors were able to achieve state-of-the-art performance on curved text benchmarks using only 10\% original polygon labels and 90\% loose or tight bounding boxes.

Wu et al. \cite{Wu2020} use text lines instead of bounding boxes as coarse labels. The labels are combined with the activation map produced by a pretrained model to filter out false positives.

\cite{Zhang} also rely on line annotations. These are used to train a model to predict text line maps, which in turn are used to filter character bounding boxes predicted by a pretrained model. Their model attains superior performance to the one by \cite{Wu2020} and is comparable to fully supervised methods.

Bonechi et al. \cite{Bonechi} use bounding box labels to generate pixel-level annotations to train a scene text segmentation model. A background-foreground segmentation network is trained based on synthetic data and then applied on the bounding box content of real datasets. 

None of these methods attempts to perform weak supervision without any sort of human labeling involved, which we aim for in this work. \\

\section{Reinforcement Learning for STD}
\label{sec:rl}
RL is an active learning paradigm which in contrast to supervised learning does not rely on labeled training data but on interactions with a predefined environment \cite{Salvador2020}. 

It is defined as a Markov Decision Process (MDP), where at each time step $t$ the so called RL agent observes a state $s_t$ in its environment and takes an action $A_t$ to manipulate the environment. During its interaction with the environment the agent receives a scalar reward according to the reward function $r$.

The agent tries to maximize the long-term reward by adapting its behavior accordingly \cite{Wyatt2015}. The agent's behavior is defined by its policy $\pi: S \longrightarrow A$ which  maps an environment state to the action the agent will take when observing that state \cite{Wyatt2015}.

The action-value function, denoted $Q^{\pi}(a_t, s_t)$, assesses the value of taking the action $a_t$ in state $s_t$ when continuing with the current policy \cite{Sutton1998}. \\

Q-learning which we use in our work is an algorithm that estimates the action-value function \cite{Wyatt2015}.

\cite{Mnih2015} introduced the concept of a \textit{deep Q-network (DQN)}, a deep neural network, which serves as a function approximator for the action-value function. In this case we speak of \textit{deep q-learning} in which the action-value function is learned using SGD \cite{Ivanov2019}.

The environment transitions are stored as tuples $(s_t, a_t, r_t, s_{t+1})$ inside a \textit{replay buffer} from which they get sampled uniformly to build batches for training \cite{Mondal2020} in order to assure independence of the training samples. 

We use the popular concept of $\epsilon$-greedy exploration \cite{Ivanov2019} for state exploration. During training actions are chosen randomly instead of being selected by the current policy with a probability controlled by the parameter $\epsilon$.

\subsection{State and Action Space}

We cast the STD problem as a MDP by letting the agent perceive the environment through a field of view restricted by a bounding box. The box initially covers the whole image. The agent can then move and resize the box until it precisely covers a text instance.

Following Caicedo et al. \cite{Caicedo2015}, the history of the last 10 taken actions are added to the state representation as well.

We also define the action set available to the agent similarly to Caicedo et al. \cite{Caicedo2015}. 8 discrete actions allow the agent to move the bounding box along the x- and y-axis as well as to change its size and aspect ratio. 

The magnitude of a bounding box movement is set proportional to the current width and height of the bounding box and controlled by a hyperparameter $\alpha$ \cite{Caicedo2015}.

The action set is completed by the \textit{trigger} action, which the agent uses to end the search once the text instance has been found.


\subsection{Reward Function}
\label{sec:reward_functions}

We design a sparse reward function, i.e. the reward is only assigned to the final \textit{trigger} action, all other actions receive a reward of $0$. 
In contrast to the approach by Caicedo et al., the reward the agent receives is based on the tightness-aware intersection-over-union (TIoU) between the detection of a word and the actual ground-truth area. The TIoU was developed by Liu et al. \cite{Liu} as an evaluation metric which penalizes detections that cut off parts of the text. To achieve this, the original IoU is weighted proportionally to the size of the text area extending outside of the detection bounding box \cite{Liu}. 

For supervised learning we define the reward $R_{\mathrm{trigger}}$ proportionally to the TIoU with the ground-truth bounding box. To give the agent an incentive to finish an episode quickly we additionally subtract a penalty proportional to the number of steps taken.

\begin{equation}
    \label{eq:tightness_reward}
    R_{\mathrm{trigger}} = \eta * TIoU(d, g) - s * p
\end{equation}

Here, $d$ and $g$ are the detection and ground-truth boxes respectively. $\eta$ and $p$ are constants and $s$ is the number of steps taken to find the text.

With weak supervision, information about how much text lies outside of the bounding box is not available, hence we use the standard IoU in this case.

To be able to detect multiple words per image the search has to be reset to the entire image after the \textit{trigger} action has been used. 
Caicedo et al. \cite{Caicedo2015} introduce the concept of inhibition of return (IoR). The environment draws a black cross, known as the IoR marker, onto the episode image, masking out the detected text area. This inhibits the word from being picked up again in the following search. The marker is drawn onto the closest ground-truth bounding box to the detection, so that the text is guaranteed to be covered completely. During testing the bounding box predicted by the agent is covered up.

\subsection{Upscaling of Ground-Truth Boxes}

In the next two sections we propose additional optimizations to the training procedure. A simple way of improving the detection accuracy is to enlarge the ground-truth boxes slightly. This leaves a thin text-free margin around words and makes it easier for the algorithm to determine if a text instance is entirely covered by the detection bounding box. 

It also biases the agent towards larger detection bounding boxes which are more likely to cover the text completely leading to a precise fit of the IoR markers. If the markers leave out parts of a word, these might get picked up in a following search producing false positives.

\subsection{Framestacking}

Additionally, we stack multiple environment frames in the policy input. This concept is known as \textit{framestacking} in the RL literature \cite{Espeholt, Shang}. It is frequently used in Atari video game environments. We follow these approaches and stack grayscale images of the four last environment states along the channel dimension. Reducing the color information to one channel keeps the processing overhead low as only 4 instead of the initial 3 channels have to be processed by the feature extractor.

The aim of this idea is that predictions become more accurate if the action is not solely decided based on one state image but multiple previous states, i.e. the trajectory that lead to the current state. This increases the information density on which the policy bases its prediction and leads to better detections. 

\section{Weak Supervision Solution}
\label{sec:weak_sup}
Now that we have introduced RL for supervised STD we want to combine it with a weakly supervised approach.
Our reward function definition solely relies on the IoU between detection and ground-truth. Instead of being calculated based on ground-truth information, the IoU can also be estimated using a neural network which we call the \textit{assessor}. 

As the \textit{assessor} can be trained using only synthetic data, this approach allows us to build a text detector that does not rely on manual ground-truth annotations. The model is based on the method by \cite{Bartz}, who propose a similar network for weakly supervised object detection.
 
 The \textit{assessor} is a neural network trained on the regression-task of predicting the IoU $iou \in \!R$ in the range $[0, 1]$ of an image crop overlapping with a text instance in an image. \\

The reward signal in the RL training can ultimately be substituted with this prediction.

The \textit{assessor}'s training data is generated by rendering words in arbitrary fonts and colors onto background images. 
Afterwards, the resulting images are randomly cropped and saved as training examples. The fact that the words are deliberately placed onto the images makes it possible to calculate the IoU, i.e. the training labels, for the crop. 


\section{Datasets}

We use the following datasets for training and evaluation: \\

\paragraph{ICDAR data} We use the test splits from the ICDAR2011 \cite{Shahab2011} and ICDAR2013 \cite{Karatzas} competition data for evaluation using the ICDAR2013 metrics. These datasets contain focused scene text where the words take up the most prominent parts of the image. ICDAR2011 contains $414$ training and $141$ test images. ICDAR2013 comes with $229$ images for training and $233$ for testing. \\
For training we expand the training splits from both dataset with images from the multi-language \textit{ICDAR2019 MLT} dataset \cite{Nayef2019} from which we extract only images containing Latin languages, adding up to $3000$ images. \\
We merge the individual datasets into a combined ICDAR training dataset consisting of $3643$ images. \\

\paragraph{SynthText} For our semi-supervision experiments we also use data from \textit{SynthText} \cite{Gupta2016}, an expansive synthetic dataset for text detection. \\
A RL training on the complete datasets containing $800k$ images would result in very long training times, which is why we extract a random $5\%$ subset, i.e., $40k$ images, from it. \\

\paragraph{Assessor Training Data} For the data on which the \textit{asseessor} model is trained we take the background images used in the \textit{SynthText} dataset, provided by \cite{SynthText}, and randomly render words onto them. As the RL agent is allowed to move the bounding box outside of the image, the training data also contains samples including transparent areas encoded in the alpha channel. We generate $10k$ training images. 

\section{Implementation Details}

We chose \textit{RLLib} \cite{Rllib} as RL framework. 
We use a replay buffer of 20k environment transitions and an $\epsilon$-greedy exploration with an $\epsilon$ decay from $1$ to $0.1$. $\epsilon$ is annealed over $3M$ timesteps for the ICDAR dataset, $5M$ for the larger \textit{SynthText} dataset.
We run trainings for $15M$ environment steps. \\
The discount factor $\gamma$ is set to $0.95$, the reward constant $\eta$ to $70$, the duration penalty $p$ to $0.03$ and the step size $\alpha$. \\
The environment images are resized to $224\times224$ pixels, which are passed through a ResNet-18 CNN for feature extraction. We chose ResNet-18 over ResNet-50 as the latter runs much slower during inference and did not yield significantly better results. Further hyper parameter tuning in the future should improve the performance using a larger CNN. 

Following \cite{Caicedo2015}, the past 10 actions taken by the agent are one-hot encoded and concatenated with the image embedding. This history vector has a dimension of $90$ ($10 \times 9$ actions). \\
The DQN following the feature extractor consists of two hidden linear layers with of $1024$ neurons ReLU activations and one output layer. \\

Following \cite{Bartz}, the \textit{assessor} model is made up of a residual CNN without bias parameters. It consists of 4 residual blocks where the first three are made up of three convolutional layers ($3\times3$, $4\times4$ and $4\times4$ kernel sizes). The last consists of two $3\times3$ convolutional layers. All convolutional layers are followed by a ReLU activation function and group normalization layer. After a global average pooling operation a linear layer with one output neuron predicts the IoU value. We use 64 filters for the first two blocks, 128 for the third and 256 for the last block.

We train the assessor alongside the RL agent, i.e. at the end of every environment episode a training batch is passed through the network. We found this procedure to lead to better performance overall compared to pretraining the assessor.

All neural networks are trained with a learning rate of $1e^{-4}$. \\

\section{Results}
\label{sec:results}
We share the results of our study separately for the supervised RL model and the model combined with the weak supervision technique. 

\subsection{Supervised Training}

In the following we compare detection results on ICDAR2011 and ICDAR2013 benchmarks by our method to those of related work. A direct comparison is only possible with the only other RL method by \cite{Wang}. This is because the application of RL to STD is still in its early stages and some optimizations present in most state-of-the-art solutions are not yet included in our approach. 

One of such optimizations is handling multi-scale detection cases. For instance, \cite{Liao2016} mention that performing multiple network passes in five different scales increased their f1-score by $5\%$.

Additionally, most other approaches use higher resolution input images, e.g. \textit{CTPN} \cite{Tian2016} scales the input images to 600 pixels on the shortest side, whereas we use $224 \times 224$ images. \textit{CCTN} \cite{He2016} uses a scale of $500 \times 500$ pixels. 

Consequently it is expected that the recent regression methods outperform our supervised training. It is more a question of how close our solution gets and to outline improvements which are needed to push the results even further.  \\

We compare against methods targeting horizontal rectangular text, namely \textit{CTPN}, \textit{TextBoxes}, \textit{DeepText} and \textit{CCTN}. Additionally, we mention the results obtained using Faster-RCNN, which can be considered a baseline as it is in its bare form developed for general object detection. 

Most importantly it can be said, that our training improvements involving the TIoU reward, framestacking and upscaling of ground-truth boxes pushed our method to achieve significantly better results than the RL solution by \cite{Wang2018}. We more than doubled their recall on ICDAR2013 (going from $32\%$ to $70\%$) and improved the precision by $4\%$. \\

Compared to Faster R-CNN\footnote{Text detection results as reported in \cite{Zhong2016}} our solution achieves a $4\%$ higher f1-score on ICDAR2013 and scores equally on ICDAR2011. For the other methods, we are $8\%$ in f1-score behind the closest competitor (\textit{TextBoxes}) on ICDAR2013. For ICDAR2011 our method scores $11\%$ lower in f1 compared to \textit{DeepText}.

\begin{table}[!t]
    \caption{Comparison of Our Method with Related Work on ICDAR2013 and ICDAR2011; P, R, F Stand for Precision, Recall and F1-score}
    \label{tab:icdar_results}
    \centering
    \renewcommand{\arraystretch}{1.2}
    \begin{tabular}{|lcccccc|}
        \hline
        \multirow{2}{*}{Method} & \multicolumn{3}{c}{ICDAR2013} & \multicolumn{3}{c|}{ICDAR2011} \\ \cline{2-4} \cline{5-7}
        & P & R & F & P & R & F  \\ \hline
        CTPN \cite{Tian2016} & \textbf{0.93} & \textbf{0.83} & \textbf{0.88} & \textbf{0.89} & 0.79 & 0.84 \\
        CCTN \cite{He2016} & 0.90 & \textbf{0.83} & 0.86 & 0.88 & 0.79 & 0.84 \\
        DeepText \cite{Zhong2016} & 0.87 & \textbf{0.83} & 0.85 & 0.85 & 0.81 & 0.83 \\ 
        TextBoxes \cite{Liao2016} & 0.88 & 0.82 & 0.85 & 0.88 & \textbf{0.83} & \textbf{0.85} \\
        Faster R-CNN \cite{Ren} & 0.75 & 0.71 & 0.73 & 0.74 & 0.71 & 0.72 \\ \hline
        Wang (RL) \cite{Wang2018} & 0.80 & 0.32 & 0.46 & - & - & - \\ \hline
        Ours & 0.84 & 0.70 & 0.77 & 0.85 & 0.62 & 0.72 \\ 
        \hline
    \end{tabular}
\end{table}

\begin{figure}
    \centering
    \begin{subfigure}[b]{0.15\textwidth}
        \centering
         \includegraphics[height=28mm, width=28mm]{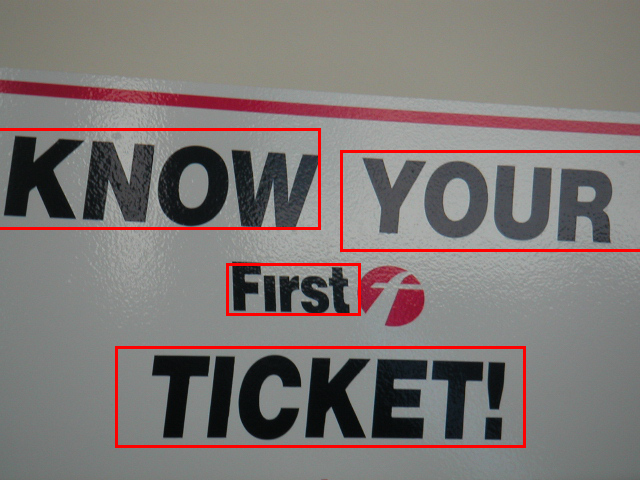}
    \end{subfigure}
    \begin{subfigure}[b]{0.15\textwidth}
        \centering
         \includegraphics[height=28mm, width=28mm]{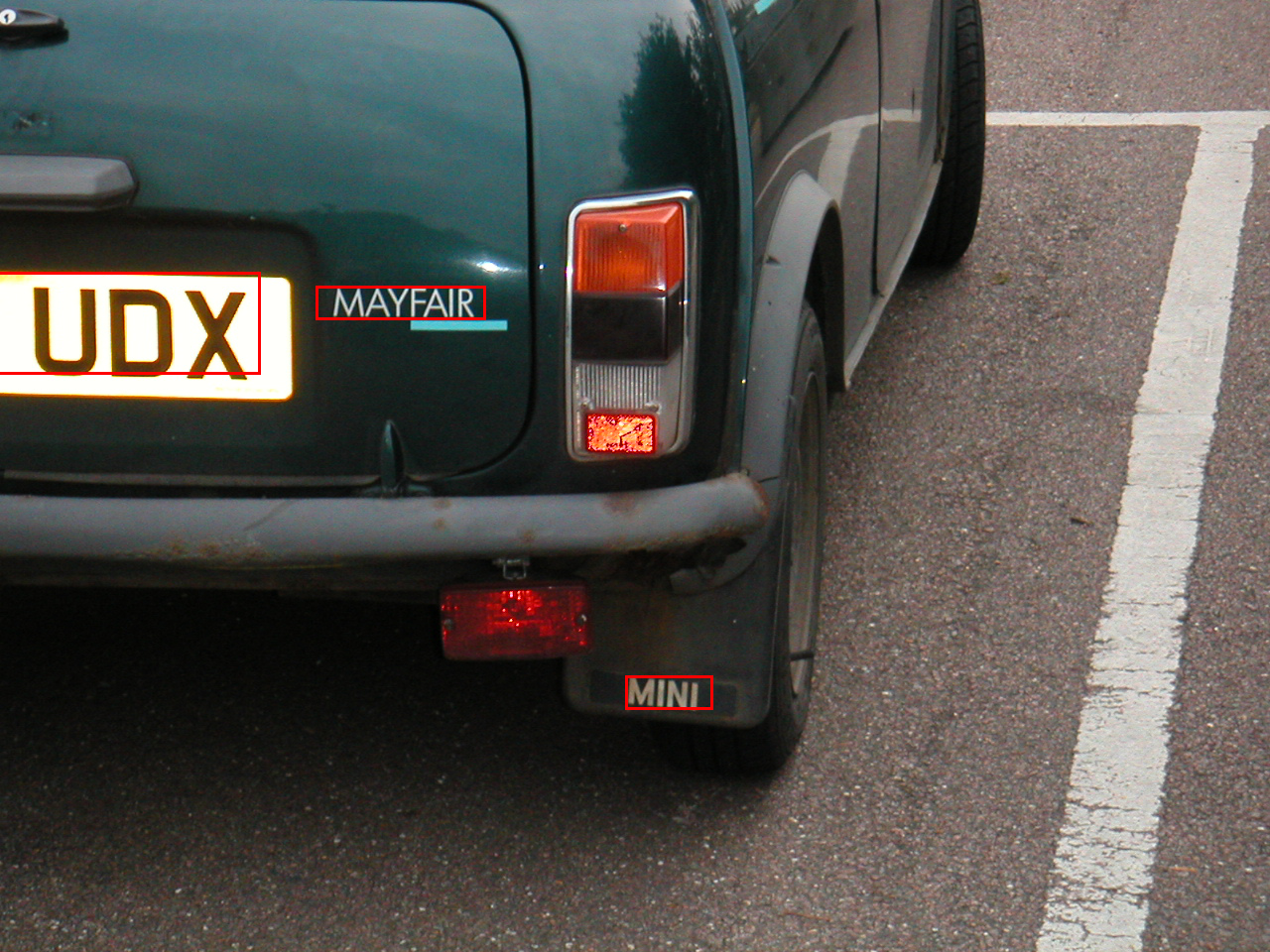}
    \end{subfigure}
    \begin{subfigure}[b]{0.15\textwidth}
        \centering
         \includegraphics[height=28mm, width=28mm]{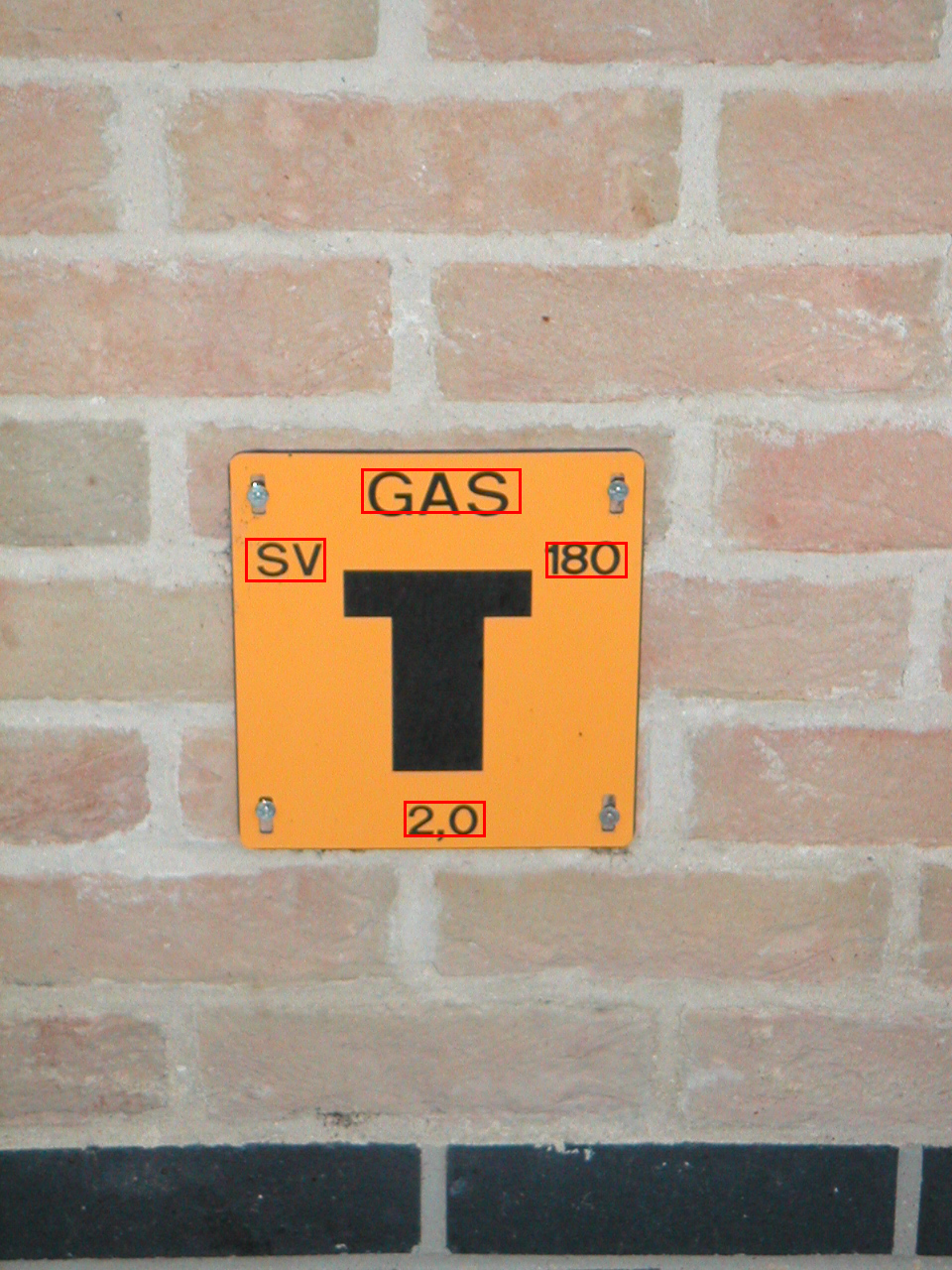}
    \end{subfigure} \\
    \vspace{0.3em}
    \begin{subfigure}[b]{0.15\textwidth}
        \centering
         \includegraphics[height=28mm, width=28mm]{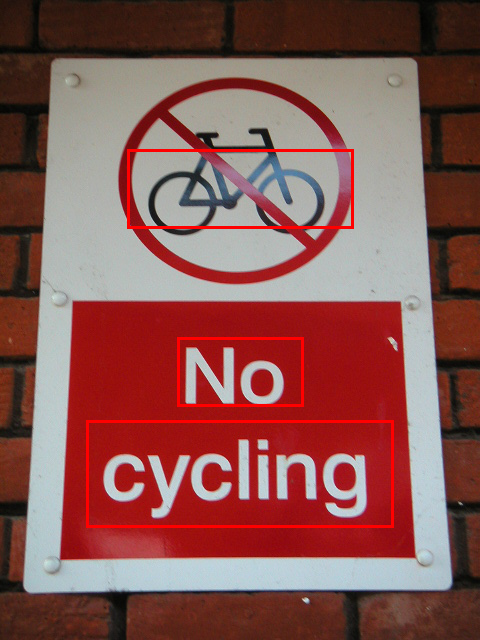}
    \end{subfigure}
    \begin{subfigure}[b]{0.15\textwidth}
        \centering
         \includegraphics[height=28mm, width=28mm]{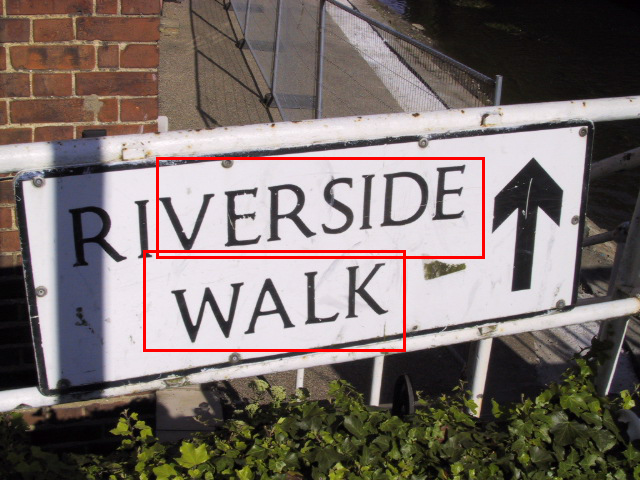}
    \end{subfigure}
    \begin{subfigure}[b]{0.15\textwidth}
        \centering
         \includegraphics[height=28mm, width=28mm]{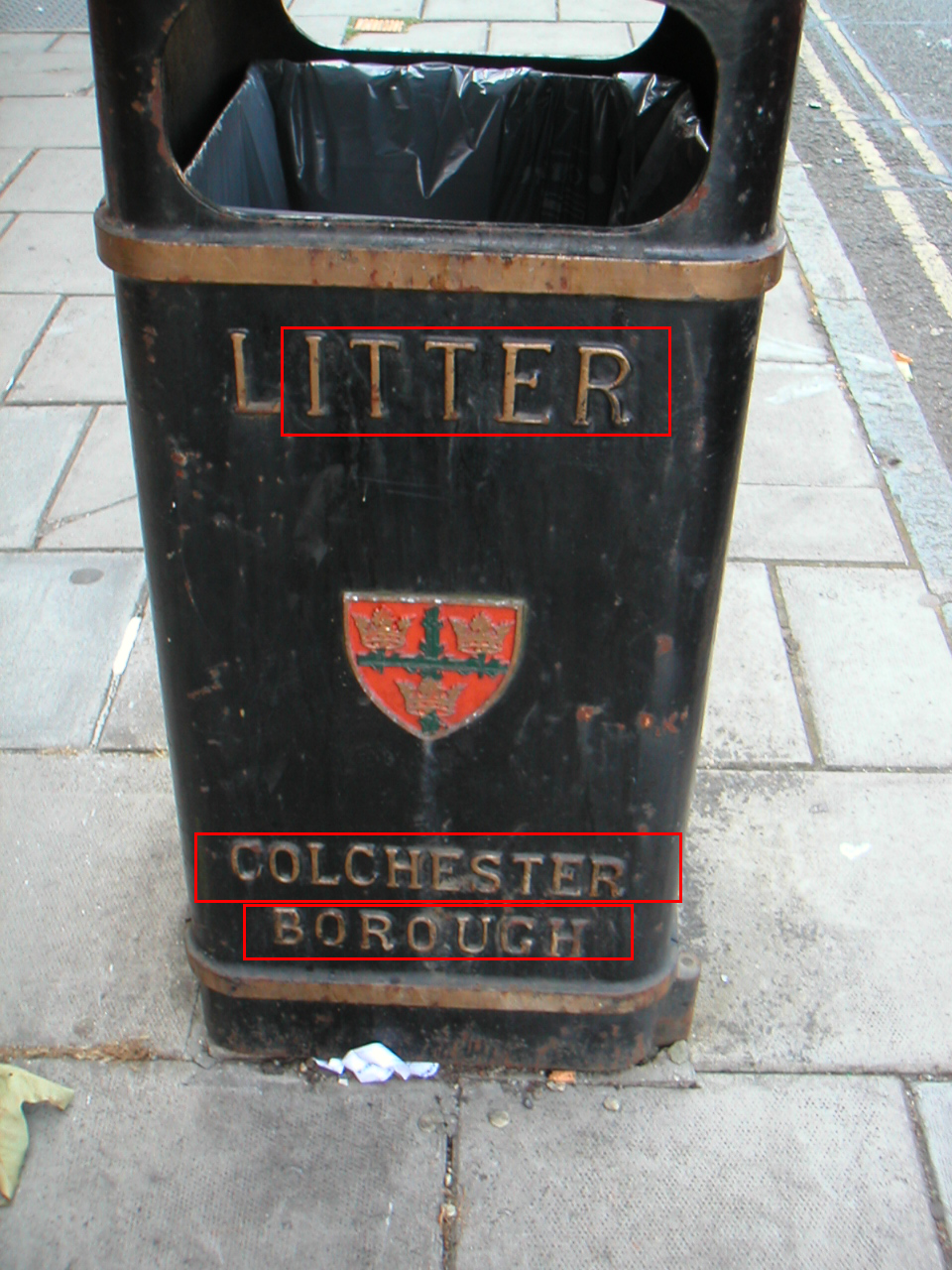}
    \end{subfigure}
    \caption{Examples of detections made by our supervised model on ICDAR2013; red rectangles are predicted bounding boxes}
    \label{fig:ic13_positive_examples}
\end{figure}

Figure \ref{fig:ic13_positive_examples} shows some qualitative results. Mainly in cases where individual words are clearly separated from each other it performs very well. Sometimes the model mistakes icons, such as the bicycle in the fourth image, for text. In longer phrases, where words appear close to each other, multiple bounding boxes are sometimes placed on single words and characters at the beginning or end of a text instance occasionally cut off. 

\subsection{Weakly Supervised Training}

\begin{figure}
    \centering
    \begin{subfigure}[b]{0.15\textwidth}
        \centering
         \includegraphics[height=28mm, width=28mm]{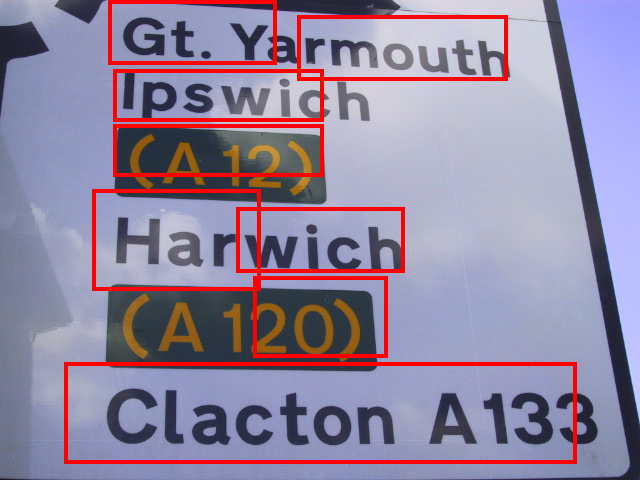}
    \end{subfigure}
    \begin{subfigure}[b]{0.15\textwidth}
        \centering
         \includegraphics[height=28mm, width=28mm]{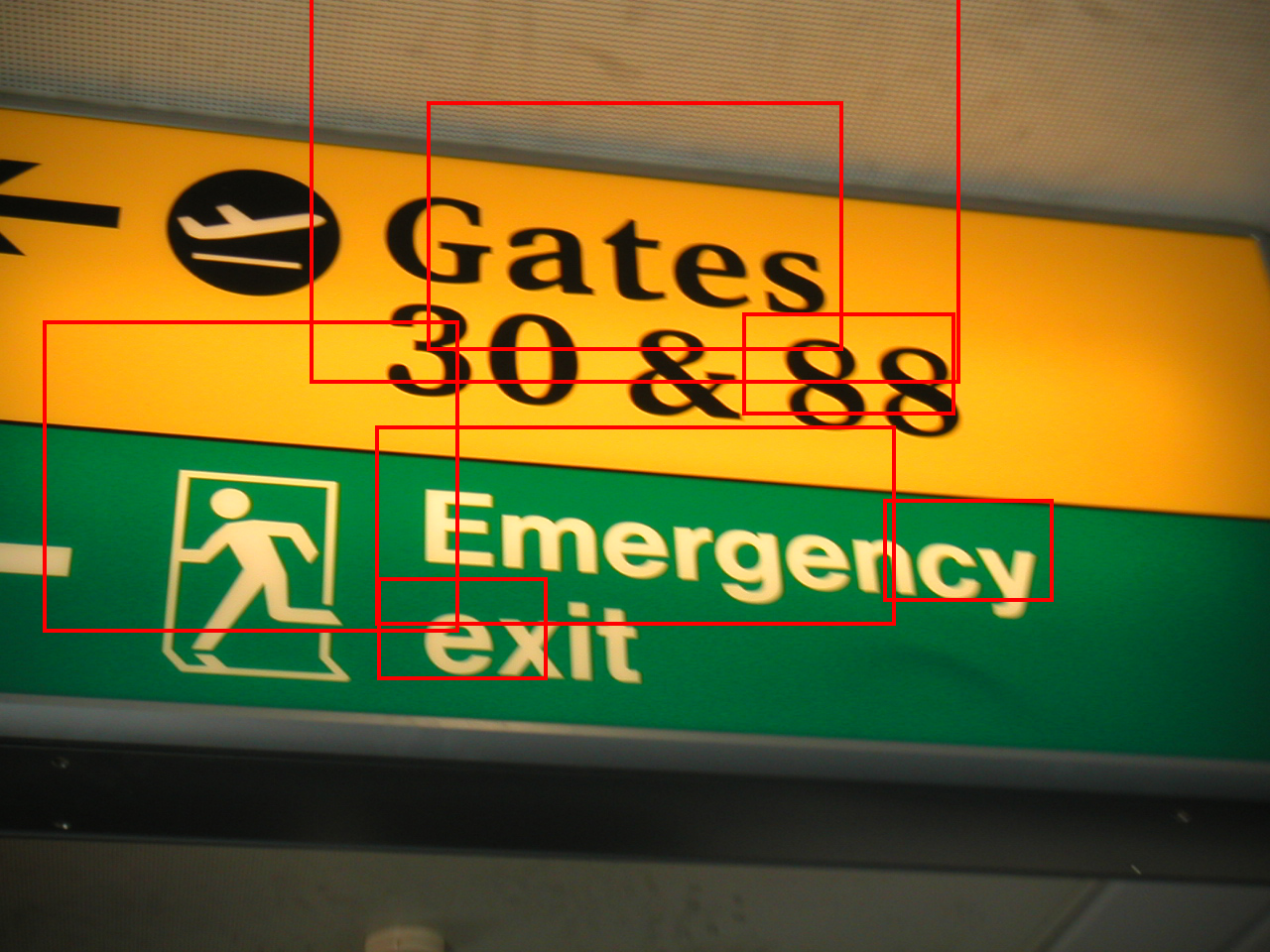}
    \end{subfigure}
    \begin{subfigure}[b]{0.15\textwidth}
        \centering
         \includegraphics[height=28mm, width=28mm]{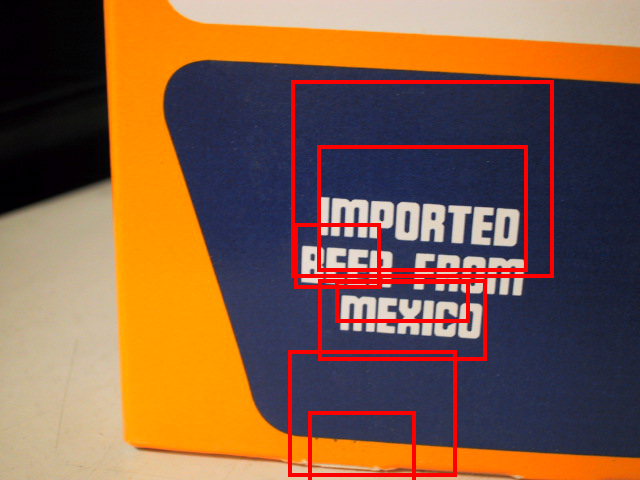}
    \end{subfigure}\\
    \vspace{0.3em}
    \begin{subfigure}[b]{0.15\textwidth}
        \centering
         \includegraphics[height=28mm, width=28mm]{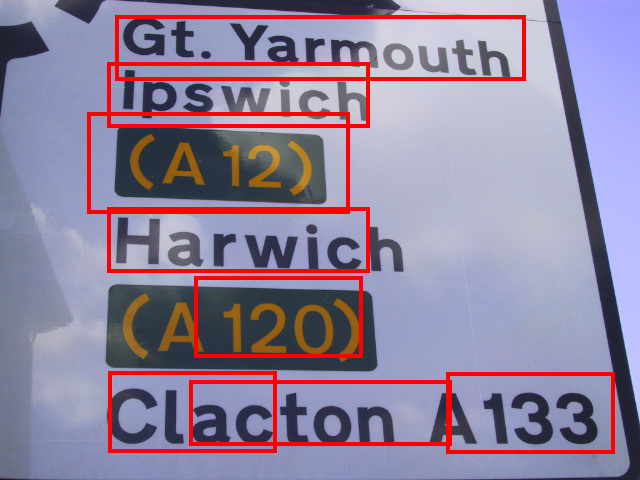}
    \end{subfigure}
    \begin{subfigure}[b]{0.15\textwidth}
        \centering
         \includegraphics[height=28mm, width=28mm]{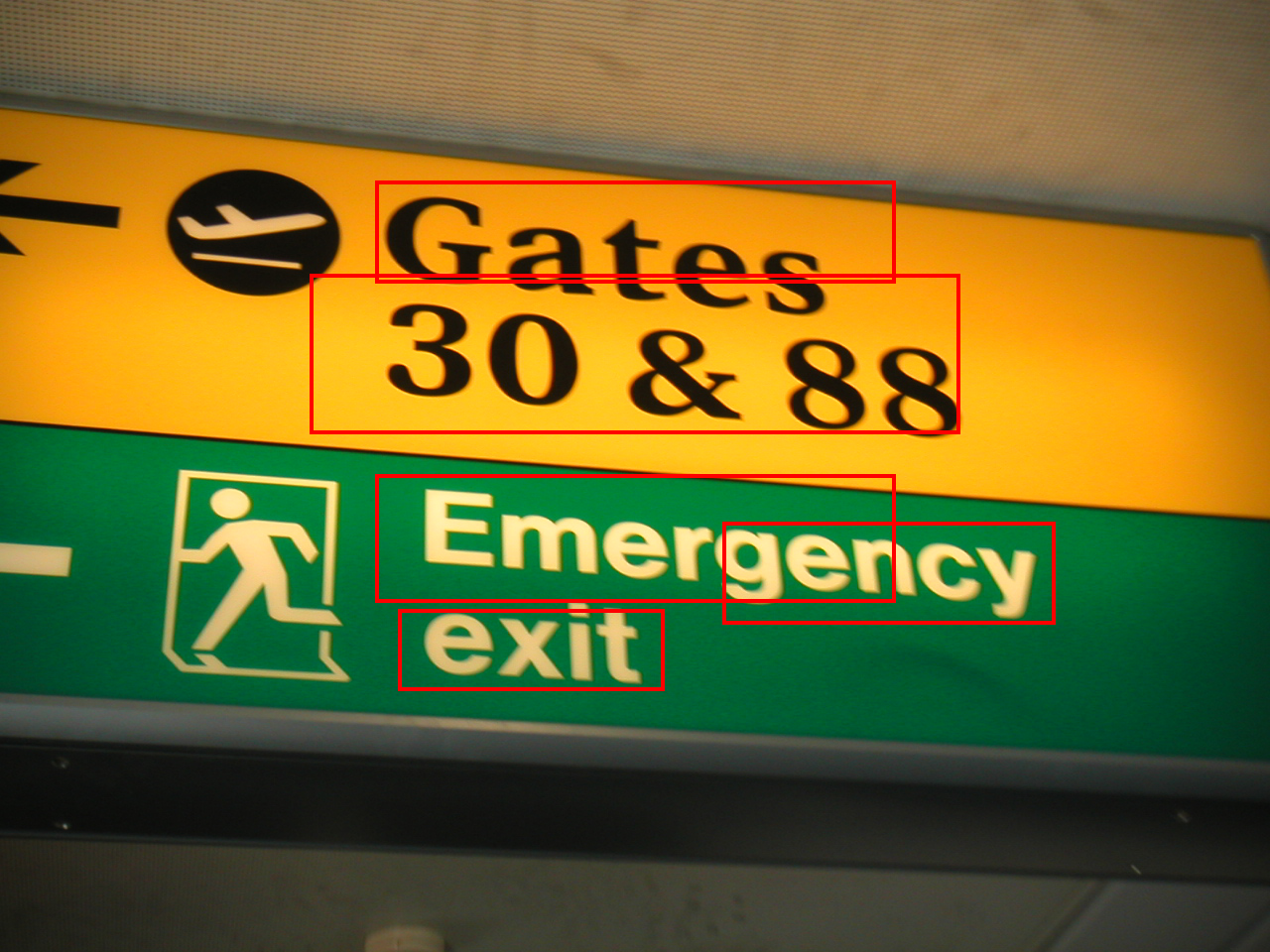}
    \end{subfigure}
    \begin{subfigure}[b]{0.15\textwidth}
        \centering
         \includegraphics[height=28mm, width=28mm]{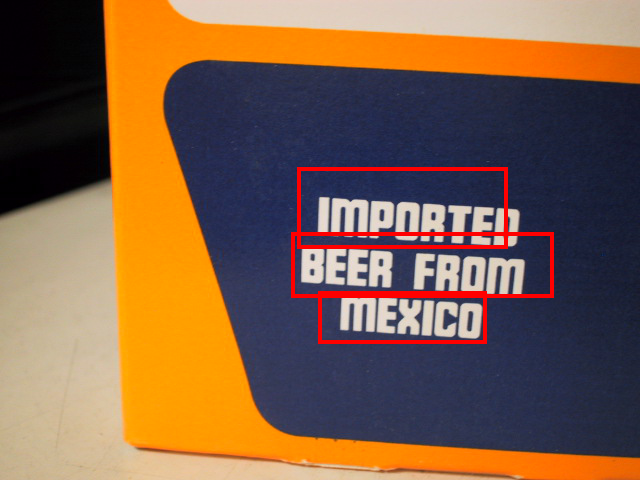}
    \end{subfigure}
    \caption{Examples of predictions on ICDAR2013 using the weakly supervised model (top row) and the semi-supervised model (bottom row); rectangles are predicted bounding boxes}
    \label{fig:assessor_examples} 
\end{figure}

Naturally, the supervised model performs much better than a model trained solely on weak labels. However, the weak supervision model achieves a recall of $36\%$. The precision on the other hand is very low with $26\%$. It is definitely possible, though, to detect text using our technique. Figure \ref{fig:assessor_examples} shows some example predictions of the weak supervision model. However, the resulting bounding boxes do lack in tightness and often do not cover the complete word. Nevertheless, the most salient text instances are clearly being detected. \\

Furthermore, we compare the weak supervision approach a semi-supervised training. For this we combine the labeled \textit{SynthText} dataset with our combined ICDAR dataset using weak labels. As \textit{SynthText} is synthetically generated, we still do not rely on any manual data annotation. \\
With this setup we attain a f1-score of $49\%$ on ICDAR2013, $13\%$ above using supervised training on \textit{SynthText} alone (see table \ref{tab:semi_supervised}). It can therefore be said that our weak supervision method proves capable of enlarging existing labeled datasets with more real-world images, which do not have do be annotated.

Figure \ref{fig:assessor_examples} shows some example detections using the semi-supervision setup. It clearly corrects many mistakes made with the solely weakly supervised model, especially reducing the number of false positive detections. 

\begin{table}[]
    \centering
    \renewcommand{\arraystretch}{1.4}
    \caption{Comparison of Supervised, Weakly Supervised and Semi-supervised training on ICDAR2013; ST Stands for SynthText, ICDAR for the combined ICDAR dataset}
    \label{tab:semi_supervised}
    \begin{tabular}{|lccc|}
        \hline
        Method & Recall & Precision & F1-Score \\ \hline
        Supervised (ST) & 0.27 & 0.56 & 0.36 \\
        Weakly supervised (ICDAR) & 0.36 & 0.26 & 0.30 \\
        Semi-supervised (ICDAR + ST) & \textbf{0.43} & \textbf{0.57} & \textbf{0.49} \\
        \hline
    \end{tabular}
    \vspace{-1.5em}
\end{table}

\section{Discussion}
\label{sec:discussion}
\begin{figure}
    \centering
    \includegraphics[width=0.4\textwidth]{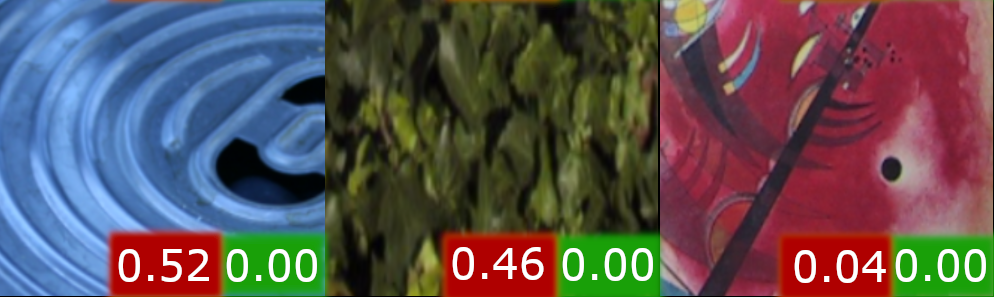}
    \caption{Examples of assessor predictions on background; IoU prediction on red square, ground-truth IoU on green square }
    \label{fig:assessor_predictions}
    \vspace{-2em}
\end{figure}

While a combination of supervised learning on synthetic data and weakly supervised learning on real-world data produced good first results, the weak supervision approach alone is still very imprecise and produces too many false positive detections. We attribute this to the assessor's sensitivity to textured backgrounds. Figure \ref{fig:assessor_predictions} shows the assessor prediction results compared to the ground-truth IoU values for some example background regions in real-world images. Whereas it correctly predicts a low IoU value for many background areas, as in the last example, it sometimes predicts rather high values (between $0.30$ and $0.60$) for regions with repeating patterns. This leads to many trigger actions by the RL agent on non-text areas.

We believe that there is still a lot of potential for improvement in the assessor design. We took the network proposed by \cite{Bartz}, but other architectures might work better to predict text IoU values. The training procedure offers plenty of room for further experimentation as well. Including more variations in the training images, finding better background images for the data generation or using existing labeled real-world data for the assessor training could lead to better learning of the model.

Our weak supervision proof-of-concept turned out to be feasible as we were able to get reasonable results without any human data annotation required. 

Concerning the supervised model there are several improvements with a lot of potential,
which we propose for future work. First of all, our feature extractor only relies on convolutional filters of fixed scales. Using multi-scale CNNs, such as a feature pyramid network \cite{Lin2016}, would probably improve the detection results further. \\

The RL training procedure also offers potential for further research. Deep q-learning, which we used for our work, is only one of many RL algorithms. Most of which have not been used in this context yet. Policy gradient methods, for instance, are known to be able to handle larger action spaces and require no tuning of the environment exploration. It is therefore worthwhile to explore their effectiveness when used together with different action sets for the agent such as with added rotation and sheering actions for bounding boxes. 

The weak supervision model is sometimes confused by strong contrast around the colored IoR markers. Other ways of hiding already detected texts are hence worthwhile to explore. Generative adversarial networks could be used to fill the respective image areas with generated background textures. 

\section{Conclusion}

We explored reinforcement learning (RL) in the context of scene text detection and used it to develop a weakly-supervised detection model. We introduced multiple training optimizations, including "framestacking" of the environment state history and the application of a tightness-aware IoU \cite{Liu} reward to the supervised RL model This way our model outperformed the related method on RL for STD by a large margin. We achieved competitive benchmark results compared to state-of-the-art regression-based algorithms.

Our weak supervision technique adapts the design of Bartz et al. \cite{Bartz} to estimate the scalar reward during RL training. The network predicts the IoU of a detection bounding box and is trained solely on synthetic data.

Training on unlabeled data using this method proved to be feasible. On real-world images it achieved an f1-score of $30\%$. Training in a semi-supervised fashion on a combination of real-world images labeled using the weak supervision technique and supervised synthetic data improved the results to reach an f1-score of $49\%$, $13\%$ better than relying on the supervised data alone. The method can therefore be considered as a tool to enlarge existing datasets with unlabeled data. \\

\bibliographystyle{IEEEtran}
\bibliography{IEEEabrv,references}

\begin{thebibliography}{10}
\providecommand{\url}[1]{#1}
\csname url@samestyle\endcsname
\providecommand{\newblock}{\relax}
\providecommand{\bibinfo}[2]{#2}
\providecommand{\BIBentrySTDinterwordspacing}{\spaceskip=0pt\relax}
\providecommand{\BIBentryALTinterwordstretchfactor}{4}
\providecommand{\BIBentryALTinterwordspacing}{\spaceskip=\fontdimen2\font plus
\BIBentryALTinterwordstretchfactor\fontdimen3\font minus
  \fontdimen4\font\relax}
\providecommand{\BIBforeignlanguage}[2]{{%
\expandafter\ifx\csname l@#1\endcsname\relax
\typeout{** WARNING: IEEEtran.bst: No hyphenation pattern has been}%
\typeout{** loaded for the language `#1'. Using the pattern for}%
\typeout{** the default language instead.}%
\else
\language=\csname l@#1\endcsname
\fi
#2}}
\providecommand{\BIBdecl}{\relax}
\BIBdecl

\bibitem{He}
K.~He, X.~Zhang, S.~Ren, and J.~Sun, ``Deep residual learning for image
  recognition,'' in \emph{2016 IEEE Conference on Computer Vision and Pattern
  Recognition (CVPR)}, 2016, pp. 770--778.

\bibitem{Redmon}
J.~Redmon, S.~Divvala, R.~Girshick, and A.~Farhadi, ``You only look once:
  Unified, real-time object detection,'' in \emph{2016 IEEE Conference on
  Computer Vision and Pattern Recognition (CVPR)}, 2016, pp. 779--788.

\bibitem{UjjalDey}
A.~Dey, S.~K. Ghosh, and E.~Valveny, ``Beyond visual semantics: Exploring the
  role of scene text in image understanding,'' \emph{Pattern Recognit. Lett.},
  vol. 149, pp. 164--171, 2021.

\bibitem{Yang}
\BIBentryALTinterwordspacing
Z.~Yang, M.~Shi, Y.~Avrithis, C.~Xu, and V.~Ferrari, ``{Training Object
  Detectors from Few Weakly-Labeled and Many Unlabeled Images}.'' [Online].
  Available: \url{https://hal.inria.fr/hal-02393688}
\BIBentrySTDinterwordspacing

\bibitem{Zhao}
\BIBentryALTinterwordspacing
M.~Zhao, W.~Feng, F.~Yin, X.-Y. Zhang, and C.-L. Liu, ``{Weakly-Supervised
  Arbitrary-Shaped Text Detection with Expectation-Maximization Algorithm},''
  2020. [Online]. Available: \url{http://arxiv.org/abs/2012.00424}
\BIBentrySTDinterwordspacing

\bibitem{Cao2020}
\BIBentryALTinterwordspacing
D.~Cao, Y.~Zhong, L.~Wang, Y.~He, and J.~Dang, ``Scene text detection in
  natural images: A review,'' \emph{Symmetry}, vol.~12, no.~12, 2020. [Online].
  Available: \url{https://www.mdpi.com/2073-8994/12/12/1956}
\BIBentrySTDinterwordspacing

\bibitem{Khan2021}
T.~Khan, R.~Sarkar, and A.~Mollah, ``Deep learning approaches to scene text
  detection: a comprehensive review,'' \emph{Artificial Intelligence Review},
  vol.~54, pp. 1--60, 06 2021.

\bibitem{Lin2020}
\BIBentryALTinterwordspacing
H.~Lin, P.~Yang, and F.~Zhang, ``{Review of Scene Text Detection and
  Recognition},'' \emph{Archives of Computational Methods in Engineering},
  vol.~27, no.~2, pp. 433--454, 2020. [Online]. Available:
  \url{https://doi.org/10.1007/s11831-019-09315-1}
\BIBentrySTDinterwordspacing

\bibitem{Wang}
Y.~Wang, H.~Xie, Z.-J. Zha, M.~Xing, Z.~Fu, and Y.~Zhang, ``Contournet: Taking
  a further step toward accurate arbitrary-shaped scene text detection,'' in
  \emph{2020 IEEE/CVF Conference on Computer Vision and Pattern Recognition
  (CVPR)}, 2020, pp. 11\,750--11\,759.

\bibitem{Lucas2005}
S.~M. Lucas, ``{ICDAR 2005 text locating competition results},''
  \emph{Proceedings of the International Conference on Document Analysis and
  Recognition, ICDAR}, vol. 2005, no. September, pp. 80--84, 2005.

\bibitem{Yin}
X.-C. Yin, X.~Yin, K.~Huang, and H.-W. Hao, ``Robust text detection in natural
  scene images,'' \emph{IEEE Transactions on Pattern Analysis and Machine
  Intelligence}, vol.~36, no.~5, pp. 970--983, 2014.

\bibitem{Kong}
X.~Kong, B.~Xin, Y.~Wang, and G.~Hua, ``{Collaborative deep reinforcement
  learning for joint object search},'' in \emph{Proceedings - 30th IEEE
  Conference on Computer Vision and Pattern Recognition, CVPR 2017}, vol.
  2017-Janua, 2017, pp. 7072--7081.

\bibitem{Caicedo2015}
J.~C. Caicedo and S.~Lazebnik, ``{Active object localization with deep
  reinforcement learning},'' in \emph{Proceedings of the IEEE International
  Conference on Computer Vision}, vol. 2015 Inter, 2015, pp. 2488--2496.

\bibitem{Bartz}
C.~Bartz, H.~Yang, J.~Bethge, and C.~Meinel, ``Loans: Weakly supervised object
  detection with localizer assessor networks,'' in \emph{ACCV Workshops}, 2018.

\bibitem{Ren}
S.~Ren, K.~He, R.~Girshick, and J.~Sun, ``Faster r-cnn: Towards real-time
  object detection with region proposal networks,'' in \emph{Advances in Neural
  Information Processing Systems}, C.~Cortes, N.~Lawrence, D.~Lee, M.~Sugiyama,
  and R.~Garnett, Eds.\hskip 1em plus 0.5em minus 0.4em\relax Curran
  Associates, Inc.

\bibitem{Ma}
J.~Ma, W.~Shao, H.~Ye, L.~Wang, H.~Wang, Y.~Zheng, and X.~Xue,
  ``Arbitrary-oriented scene text detection via rotation proposals,''
  \emph{IEEE Transactions on Multimedia}, vol.~20, no.~11, pp. 3111--3122,
  2018.

\bibitem{Zhong2016}
\BIBentryALTinterwordspacing
Z.~Zhong, L.~Jin, S.~Zhang, and Z.~Feng, ``Deeptext: {A} unified framework for
  text proposal generation and text detection in natural images,'' \emph{CoRR},
  vol. abs/1605.07314, 2016. [Online]. Available:
  \url{http://arxiv.org/abs/1605.07314}
\BIBentrySTDinterwordspacing

\bibitem{Jiang2017}
\BIBentryALTinterwordspacing
Y.~Jiang, X.~Zhu, X.~Wang, S.~Yang, W.~Li, H.~Wang, P.~Fu, and Z.~Luo,
  ``{R2CNN: Rotational Region CNN for Orientation Robust Scene Text
  Detection},'' \emph{arXiv}, jun 2017. [Online]. Available:
  \url{http://arxiv.org/abs/1706.09579}
\BIBentrySTDinterwordspacing

\bibitem{Tian2016}
\BIBentryALTinterwordspacing
Z.~Tian, W.~Huang, T.~He, P.~He, and Y.~Qiao, ``{Detecting Text in Natural
  Image with Connectionist Text Proposal Network},'' \emph{Lecture Notes in
  Computer Science (including subseries Lecture Notes in Artificial
  Intelligence and Lecture Notes in Bioinformatics)}, vol. 9912 LNCS, pp.
  56--72, sep 2016. [Online]. Available:
  \url{https://arxiv.org/abs/1609.03605v1}
\BIBentrySTDinterwordspacing

\bibitem{He2016}
\BIBentryALTinterwordspacing
T.~He, W.~Huang, Y.~Qiao, and J.~Yao, ``Accurate text localization in natural
  image with cascaded convolutional text network,'' \emph{CoRR}, vol.
  abs/1603.09423, 2016. [Online]. Available:
  \url{http://arxiv.org/abs/1603.09423}
\BIBentrySTDinterwordspacing

\bibitem{Long2018}
\BIBentryALTinterwordspacing
S.~Long, J.~Ruan, W.~Zhang, X.~He, W.~Wu, and C.~Yao, ``{TextSnake: A Flexible
  Representation for Detecting Text of Arbitrary Shapes},'' \emph{Lecture Notes
  in Computer Science (including subseries Lecture Notes in Artificial
  Intelligence and Lecture Notes in Bioinformatics)}, vol. 11206 LNCS, pp.
  19--35, jul 2018. [Online]. Available:
  \url{https://arxiv.org/abs/1807.01544v2}
\BIBentrySTDinterwordspacing

\bibitem{Wangb}
W.~Wang, E.~Xie, X.~Li, W.~Hou, T.~Lu, G.~Yu, and S.~Shao, ``Shape robust text
  detection with progressive scale expansion network,'' in \emph{2019 IEEE/CVF
  Conference on Computer Vision and Pattern Recognition (CVPR)}, 2019, pp.
  9328--9337.

\bibitem{Wang2018}
H.~Wang, S.~Huang, and L.~Jin, ``{Focus on Scene Text Using Deep Reinforcement
  Learning},'' \emph{Proceedings - International Conference on Pattern
  Recognition}, vol. 2018-Augus, pp. 3759--3765, 2018.

\bibitem{Mathe}
S.~Mathe, A.~Pirinen, and C.~Sminchisescu, ``{Reinforcement Learning for Visual
  Object Detection},'' in \emph{Proceedings of the IEEE Computer Society
  Conference on Computer Vision and Pattern Recognition}, vol. 2016-Decem,
  2016, pp. 2894--2902.

\bibitem{Bellver2016}
\BIBentryALTinterwordspacing
M.~B. Bueno, X.~Giro-i Nieto, F.~Marques, and J.~Torres, ``{Hierarchical object
  detection with deep reinforcement learning},'' in \emph{Deep Learning for
  Image Processing Applications}.\hskip 1em plus 0.5em minus 0.4em\relax IOS
  Press, nov 2017, pp. 164--176. [Online]. Available:
  \url{http://arxiv.org/abs/1611.03718}
\BIBentrySTDinterwordspacing

\bibitem{Jie}
Z.~Jie, X.~Liang, J.~Feng, X.~Jin, W.~F. Lu, and S.~Yan, ``Tree-structured
  reinforcement learning for sequential object localization,'' in
  \emph{Proceedings of the 30th International Conference on Neural Information
  Processing Systems}, ser. NIPS'16.\hskip 1em plus 0.5em minus 0.4em\relax Red
  Hook, NY, USA: Curran Associates Inc., 2016, p. 127–135.

\bibitem{Zhuang2021}
Z.~Zhuang, Z.~Liu, K.-M. Lam, S.~Huang, and G.~Dai, ``A new semi-automatic
  annotation model via semantic boundary estimation for scene text detection,''
  in \emph{Document Analysis and Recognition -- ICDAR 2021}, J.~Llad{\'o}s,
  D.~Lopresti, and S.~Uchida, Eds.\hskip 1em plus 0.5em minus 0.4em\relax Cham:
  Springer International Publishing, 2021, pp. 257--273.

\bibitem{Wu2020}
W.~Wu, J.~Xing, C.~Yang, Y.~Wang, and H.~Zhou, ``{Texts as Lines: Text
  Detection with Weak Supervision},'' \emph{Mathematical Problems in
  Engineering}, vol. 2020, 2020.

\bibitem{Zhang}
W.~Zhang, Y.~Qiu, M.~Liao, R.~Zhang, X.~Wei, and X.~Bai, ``Scene text detection
  with scribble line,'' in \emph{Document Analysis and Recognition -- ICDAR
  2021}, J.~Llad{\'o}s, D.~Lopresti, and S.~Uchida, Eds.\hskip 1em plus 0.5em
  minus 0.4em\relax Cham: Springer International Publishing, 2021, pp. 79--94.

\bibitem{Bonechi}
\BIBentryALTinterwordspacing
S.~Bonechi, M.~Bianchini, F.~Scarselli, and P.~Andreini, ``Weak supervision for
  generating pixel–level annotations in scene text segmentation,''
  \emph{Pattern Recognition Letters}, vol. 138, pp. 1--7, 2020. [Online].
  Available:
  \url{https://www.sciencedirect.com/science/article/pii/S0167865520302415}
\BIBentrySTDinterwordspacing

\bibitem{Salvador2020}
J.~Salvador, J.~Oliveira, and M.~Breternitz, ``{Reinforcement Learning: A
  Literature Review (September 2020)},'' \emph{arXiv}, no. December, pp. 1--36,
  2020.

\bibitem{Wyatt2015}
J.~Wyatt and I.~Stamatescu, \emph{\BIBforeignlanguage{English}{Reinforcement
  Learning: a brief overview}}, Jan. 2002.

\bibitem{Sutton1998}
\BIBentryALTinterwordspacing
R.~S. Sutton and A.~G. Barto, \emph{{Reinforcement Learning: An Introduction}},
  2nd~ed.\hskip 1em plus 0.5em minus 0.4em\relax The MIT Press, 2018. [Online].
  Available: \url{http://incompleteideas.net/book/the-book-2nd.html}
\BIBentrySTDinterwordspacing

\bibitem{Mnih2015}
\BIBentryALTinterwordspacing
V.~Mnih, K.~Kavukcuoglu, D.~Silver, A.~A. Rusu, J.~Veness, M.~G. Bellemare,
  A.~Graves, M.~Riedmiller, A.~K. Fidjeland, G.~Ostrovski, S.~Petersen,
  C.~Beattie, A.~Sadik, I.~Antonoglou, H.~King, D.~Kumaran, D.~Wierstra,
  S.~Legg, and D.~Hassabis, ``{Human-level control through deep reinforcement
  learning},'' \emph{Nature}, vol. 518, no. 7540, pp. 529--533, feb 2015.
  [Online]. Available: \url{https://www.nature.com/articles/nature14236}
\BIBentrySTDinterwordspacing

\bibitem{Ivanov2019}
\BIBentryALTinterwordspacing
S.~Ivanov and A.~D'yakonov, ``{Modern Deep Reinforcement Learning
  Algorithms},'' 2019. [Online]. Available:
  \url{http://arxiv.org/abs/1906.10025}
\BIBentrySTDinterwordspacing

\bibitem{Mondal2020}
A.~K. Mondal, ``A survey of reinforcement learning techniques: Strategies,
  recent development, and future directions,'' \emph{ArXiv}, vol.
  abs/2001.06921, 2020.

\bibitem{Liu}
Y.~Liu, L.~Jin, Z.~Xie, C.~Luo, S.~Zhang, and L.~Xie, ``Tightness-aware
  evaluation protocol for scene text detection,'' in \emph{2019 IEEE/CVF
  Conference on Computer Vision and Pattern Recognition (CVPR)}, 2019, pp.
  9604--9612.

\bibitem{Espeholt}
\BIBentryALTinterwordspacing
L.~Espeholt, H.~Soyer, R.~Munos, K.~Simonyan, V.~Mnih, T.~Ward, Y.~Doron,
  V.~Firoiu, T.~Harley, I.~Dunning, S.~Legg, and K.~Kavukcuoglu, ``{IMPALA:}
  scalable distributed deep-rl with importance weighted actor-learner
  architectures,'' in \emph{Proceedings of the 35th International Conference on
  Machine Learning, {ICML} 2018, Stockholmsm{\"{a}}ssan, Stockholm, Sweden,
  July 10-15, 2018}, ser. Proceedings of Machine Learning Research, J.~G. Dy
  and A.~Krause, Eds., vol.~80.\hskip 1em plus 0.5em minus 0.4em\relax {PMLR},
  2018, pp. 1406--1415. [Online]. Available:
  \url{http://proceedings.mlr.press/v80/espeholt18a.html}
\BIBentrySTDinterwordspacing

\bibitem{Shang}
\BIBentryALTinterwordspacing
W.~Shang, X.~Wang, A.~Srinivas, A.~Rajeswaran, Y.~Gao, P.~Abbeel, and
  M.~Laskin, ``Reinforcement learning with latent flow,'' \emph{CoRR}, vol.
  abs/2101.01857, 2021. [Online]. Available:
  \url{https://arxiv.org/abs/2101.01857}
\BIBentrySTDinterwordspacing

\bibitem{Shahab2011}
A.~Shahab, F.~Shafait, and A.~Dengel, ``{ICDAR 2011 robust reading competition
  challenge 2: Reading text in scene images},'' in \emph{Proceedings of the
  International Conference on Document Analysis and Recognition, ICDAR}, 2011,
  pp. 1491--1496.

\bibitem{Karatzas}
D.~Karatzas, F.~Shafait, S.~Uchida, M.~Iwamura, L.~G.~i. Bigorda, S.~R. Mestre,
  J.~Mas, D.~F. Mota, J.~A. Almazàn, and L.~P. de~las Heras, ``Icdar 2013
  robust reading competition,'' in \emph{2013 12th International Conference on
  Document Analysis and Recognition}, 2013, pp. 1484--1493.

\bibitem{Nayef2019}
\BIBentryALTinterwordspacing
N.~Nayef, Y.~Patel, M.~Busta, P.~N. Chowdhury, D.~Karatzas, W.~Khlif, J.~Matas,
  U.~Pal, J.-C. Burie, C.-l. Liu, and J.-M. Ogier, ``{ICDAR2019 Robust Reading
  Challenge on Multi-lingual Scene Text Detection and Recognition --
  RRC-MLT-2019},'' \emph{Proceedings of the International Conference on
  Document Analysis and Recognition, ICDAR}, pp. 1582--1587, jul 2019.
  [Online]. Available: \url{http://arxiv.org/abs/1907.00945}
\BIBentrySTDinterwordspacing

\bibitem{Gupta2016}
\BIBentryALTinterwordspacing
A.~Gupta, A.~Vedaldi, and A.~Zisserman, ``{Synthetic Data for Text Localisation
  in Natural Images},'' \emph{Proceedings of the IEEE Computer Society
  Conference on Computer Vision and Pattern Recognition}, vol. 2016-Decem, pp.
  2315--2324, apr 2016. [Online]. Available:
  \url{http://arxiv.org/abs/1604.06646}
\BIBentrySTDinterwordspacing

\bibitem{SynthText}
\BIBentryALTinterwordspacing
SynthText, ``{ankush-me/SynthText: Code for generating synthetic text images as
  described in "Synthetic Data for Text Localisation in Natural Images", Ankush
  Gupta, Andrea Vedaldi, Andrew Zisserman, CVPR 2016.}'' [Online]. Available:
  \url{https://github.com/ankush-me/SynthText/tree/master}
\BIBentrySTDinterwordspacing

\bibitem{Rllib}
\BIBentryALTinterwordspacing
Rllib, ``{RLlib: Scalable Reinforcement Learning — Ray v2.0.0.dev0}.''
  [Online]. Available: \url{https://docs.ray.io/en/master/rllib.html}
\BIBentrySTDinterwordspacing

\bibitem{Liao2016}
\BIBentryALTinterwordspacing
M.~Liao, B.~Shi, X.~Bai, X.~Wang, and W.~Liu, ``{TextBoxes: A Fast Text
  Detector with a Single Deep Neural Network},'' \emph{31st AAAI Conference on
  Artificial Intelligence, AAAI 2017}, pp. 4161--4167, nov 2016. [Online].
  Available: \url{https://arxiv.org/abs/1611.06779v1}
\BIBentrySTDinterwordspacing

\bibitem{Lin2016}
\BIBentryALTinterwordspacing
T.-Y. Lin, P.~Doll{\'{a}}r, R.~Girshick, K.~He, B.~Hariharan, and S.~Belongie,
  ``{Feature Pyramid Networks for Object Detection},'' dec 2016. [Online].
  Available: \url{https://arxiv.org/abs/1612.03144v2}
\BIBentrySTDinterwordspacing

\end{thebibliography}

\end{document}